\documentclass[conference]{IEEEtran}
\IEEEoverridecommandlockouts
\usepackage{cite}
\usepackage{amsmath,amssymb,amsfonts}
\usepackage{algorithmic}
\usepackage{graphicx}
\usepackage{textcomp}
\usepackage{multirow}
\usepackage{hyperref}

\usepackage{xcolor}
\def\BibTeX{{\rm B\kern-.05em{\sc i\kern-.025em b}\kern-.08em
    T\kern-.1667em\lower.7ex\hbox{E}\kern-.125emX}}
\begin{document}

\title{Co-SemDepth: Fast Joint Semantic
Segmentation and Depth Estimation on Aerial Images\\

}


\author{\IEEEauthorblockN{
Yara AlaaEldin}
\IEEEauthorblockA{
\textit{University of Genova}\\
Genova, Italy\\
yara.ala96@gmail.com}
\and
\IEEEauthorblockN{
Francesca Odone}
\IEEEauthorblockA{
\textit{University of Genova}\\
Genova, Italy \\
francesca.odone@unige.it}
}

\maketitle

\begin{abstract}
Understanding the geometric and semantic properties of the scene is crucial in autonomous navigation and particularly challenging in the case of Unmanned Aerial Vehicle (UAV) navigation. Such information may be by obtained by estimating depth and semantic segmentation maps of the surrounding environment and for their practical use in autonomous navigation, 
the procedure must be performed as close to real-time as possible.
In this paper, we leverage monocular cameras on aerial robots to predict depth and semantic maps in low-altitude unstructured environments. We propose a joint deep-learning architecture that can perform the two tasks accurately and rapidly, and validate its effectiveness on MidAir and Aeroscapes benchmark datasets. 
Our joint-architecture proves to be competitive or superior to the other single and joint architecture methods while performing its task fast predicting 20.2 FPS on a single NVIDIA quadro p5000 GPU and it has a low memory footprint. All codes for training and prediction can be found on this link: \url{https://github.com/Malga-Vision/Co-SemDepth}. 
\end{abstract}

\section{Introduction}
\label{sec:intro}

The applications of aerial robotics, also known as Unmanned Aerial Vehicles (UAVs), are rapidly expanding across various fields, including environmental exploration, national security, package delivery, firefighting, and more.
As it commonly happens in autonomous navigation, sensors are adopted to estimate scene depth and semantic information. Unlike ground autonomous vehicles, many types of UAVs, including drones, have limited computational capability and allowed carried weight. Thus, not all types of sensors can be mounted on the drone, for instance LiDAR and RADAR cannot be adopted for depth estimation as they are heavy and power consuming. { Also, while LiDAR point clouds contain accurate depth information, they lack semantic meaning. To associate such depth points to their semantic meaning, an additional step of calibration between LiDAR and RGB camera  has to be done to estimate their relative transformation~\cite{velas2014calibration} and associate the points in the point cloud to their corresponding pixels in the image frame. However, this calibration is never fully accurate and this leads to errors in the semantic association.}
Other depth sensors like stereo cameras are not common and may be not appropriate for UAVs since the small baseline distance between the two internal cameras, compared to the large distance between the stereo camera and the scene, produces inaccurate depth estimates~\cite{olson2010wide}. 
Therefore, UAV applications often rely on  monocular cameras, as they are cheap, light, small in size, and can produce implicitly image pairs with large baselines, by considering non adjacent frames
~\cite{manydepth, bhat2021adabins, monodepth}. 
Video cameras also have the added value of associating to each point a semantic meaning, thanks to the availability of semantic segmentation approaches~\cite{zhao2018icnet, romera2017erfnet, xie2021segformer, yu2018bisenet, wang2018understanding}.
%


\begin{figure*}%
    \centering
    {\includegraphics[width=0.8\linewidth, height=7.2cm]{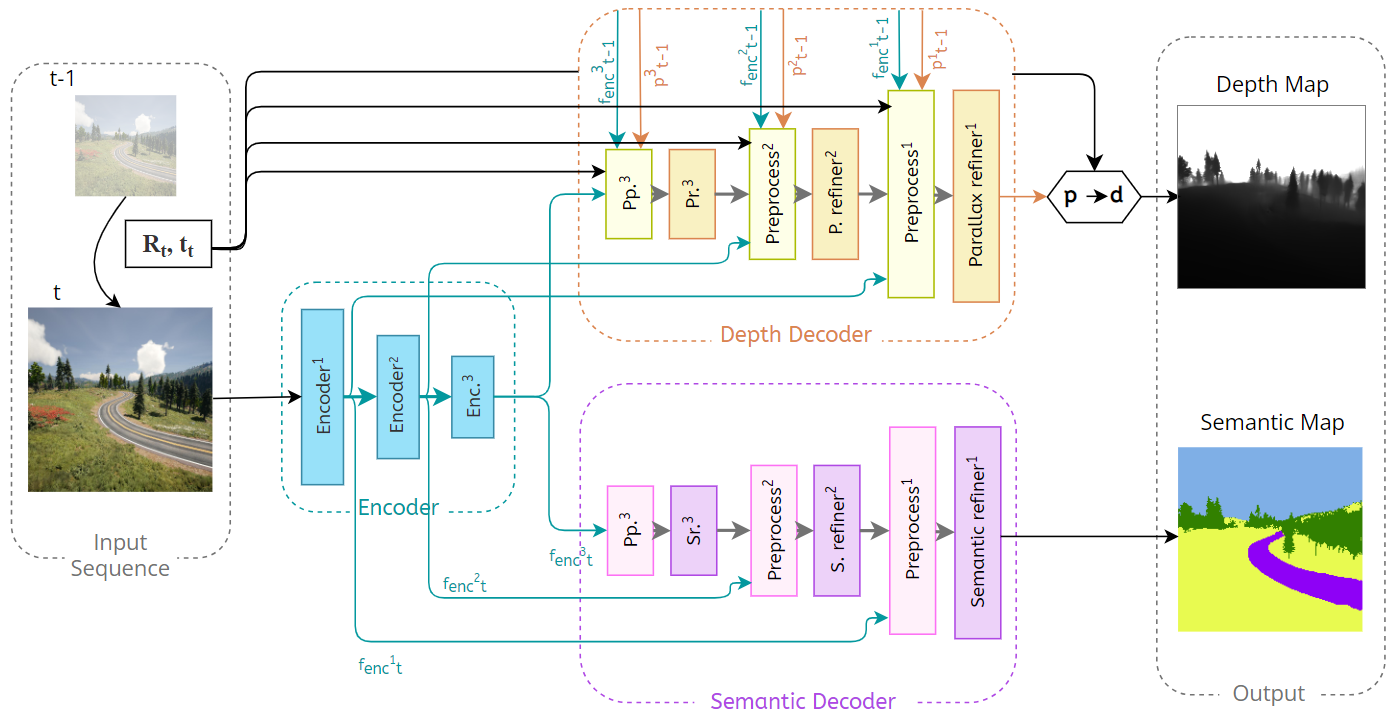} }
    \caption{\centering Our proposed Co-SemDepth Architecture. It is composed of a shared encoder and two decoders. The encoder and the depth decoder are the same presented in~\cite{m4depth}. The semantic decoder makes use of the encoded feature maps to give an estimate of the semantic segmentation map. The depth and semantic maps get scaled up as they go forward through the successive levels of the decoders. The number of shown levels here are 3 while in our experiments we use 5 levels.}%
    \label{fig:joint}%
\end{figure*}

In this paper, we leverage monocular cameras on aerial vehicles for obtaining two types of necessary information for scene understanding, {\em depth estimation} and {\em semantic segmentation}:
\begin{itemize}
    \item  In monocular depth estimation (MDE), the goal is to predict the depth of each pixel in each RGB frame captured by a video camera. Such depth expresses the distance (in meters) of the points in the world appearing in the pixels with respect to the camera frame. 
    \item In semantic segmentation, the goal is instead  to predict the semantic class of each pixel in the input RGB frames. This semantic class, expressed with a unique integer, belongs to a set of predefined semantic classes of interest that the neural network was trained on.
\end{itemize}

The two modules are complementary since depth estimation expresses the geometric properties of the scene while semantic segmentation expresses the semantic properties. We propose a joint deep architecture for achieving the two tasks accurately and in real-time, see Figure ~\ref{fig:joint} for an overview of our architecture. 
Using a {\em joint architecture} helps in saving computational time compared to performing each task separately as well as saving GPU memory by having fewer model parameters. Also, it can help in sharing learnt features between the two modules and this can in turn benefit both of them. 


Our main contributions can be summarized as the following:
\begin{itemize}
    \item We first develop a fast single architecture for semantic segmentation inspired by the MDE architecture developed in~\cite{m4depth}. We call our architecture M4Semantic and validate its effectiveness on aerial data.
    \item We merge the two architectures (M4Depth and M4Semantic) into a joint one by sharing the feature extraction part and separating the decoders. We call our joint architecture Co-SemDepth and validate its accuracy and real-time performance on aerial data.
    \item We provide a benchmark on MidAir dataset~\cite{midair} for our method and other state-of-the-art methods in semantic segmentation and depth estimation and highlight the advantages of using our joint architecture compared to the others. 
\end{itemize}
\section{Related Works}
\label{sec:related}

{ Much of the developed work in outdoor scene analysis has been driven by advancements in the automotive field, leaving aerial scene understanding comparatively under-investigated. In this section we discuss relevant state of the art, mostly referring to methods related to the automotive field.}
\subsection{Monocular Depth Estimation}
 MDE  can be addressed by means of classical methods like Structure from Motion (SfM)~\cite{sfm} or through Deep Learning methods. SfM methods generally suffer from limited accuracy and scarce availability of feature correspondences~\cite{sfm_problem}. Deep learning methods have been a way out of these low-vision challences, where the input to the DL architecture can be the whole {\em (single) image or image sequences}. Estimating depth from {single images} causes the scale ambiguity problem~\cite{scale_amb}. Some of the state of the art methods in this family are MonoDepth ~\cite{monodepth}, MonoDepth2 ~\cite{monodepth2}, Adabins ~\cite{bhat2021adabins}, and DPT ~\cite{vtd}. Estimating depth from {image sequences} improves the depth estimates compared to single image methods by making use of the temporal relation between input video frames~\cite{patil2020don, manydepth, m4depth, kopf2021robust} to better understand the scale of objects. 

The training of MDE networks can be realized in a {\em supervised or self-supervised fashion}. In {supervised methods}~\cite{li2017two,laina2016deeper,eigen2014depth,cs2018depthnet,m4depth}, depth maps ground truth should be provided to the network. The main challenge for these methods is the scarcity of available annotated datasets, especially in the aerial domain~\cite{midair, tartanair,wilduav}, that should cover different scenarios and environments. Some of the most used datasets in the literature are KITTI~\cite{kitti} for autonomous driving applications, and NYU~\cite{nyu} for indoor scenes. {Self-supervised methods} can be trained to predict the depth without requiring ground truth depth maps. They use as a self-supervision the nearby video frames ~\cite{casser2019depth,manydepth, masoumian2023gcndepth} or synchronized stereo pairs ~\cite{huang2019unsupervised,monodepth}. 

\textbf{Open problems:} Few MDE methods mentioned their inference time and no benchmarking was found in the literature comparing the inference time of different MDE methods. Very few methods were benchmarked on low-altitude aerial datasets~\cite{m4depth,miclea2021monocular}.
We address such challenges by benchmarking our method on the aerial dataset MidAir~\cite{midair}. In this benchmark we use both depth evaluation metrics and inference time to compare our performance with other state-of-the-art.

\subsection{Semantic Segmentation}
Semantic segmentation  is widely addressed in the literature, in particular in the autonomous driving field~\cite{zhao2018icnet, romera2017erfnet, xie2021segformer, yu2018bisenet, wang2018understanding}  using CityScapes~\cite{cordts2016cityscapes} dataset for benchmarking.

Unlike depth estimation, semantic segmentation networks are normally trained in a supervised fashion. This can be realized using image-based or video-based methods. Some state-of-the-art examples of {\em image-based methods} are PSPNet~\cite{zhao2017pyramid}, ICNet~\cite{zhao2018icnet}, ERFNet~\cite{romera2017erfnet}, Bisenet~\cite{yu2018bisenet}, and SegFormer~\cite{xie2021segformer}. Many {\em video-based methods}~\cite{jain2019accel, xu2018dynamic, paul2020efficient, mahasseni2017budget, li2021video, hu2020temporally, liu2020efficient, sun2022coarse} make benefit from the fact that there is a big overlap  between successive  frames by avoiding to repeat the semantic segmentation computation on each frame. They extract features of key frames and use feature propagation using light optical flow networks or interpolation techniques to pass the information to sequential video frames. This is done to speed-up inference time. However, the mIoU achieved using these methods is usually lower than image-based methods.

\textbf{Open problems:} few semantic segmentation methods~\cite{nedevschi2021weakly, aeroscapes, zheng2023deep} were benchmarked on low-altitude aerial datasets 
%
We benchmark our method on MidAir dataset for semantic segmentation. We also adapt the codes of other state-of-the-art methods and benchmark them on MidAir. 
In addition, we benchmark our method on Aeroscapes~\cite{aeroscapes} dataset. 

\subsection{Joint Architectures}
The idea of joint or multi-task deep learning architectures has been explored in the literature for various vision tasks. In~\cite{xu2020multi}, a joint architecture is implemented for image segmentation and classification, while in~\cite{qin2018joint} they developed a joint network for motion estimation and segmentation. There are some works~\cite{cao2016exploiting, mousavian2016joint, nekrasov2019real, liu2018collaborative, zhang2018joint, he2021sosd} that addressed joint depth estimation and semantic segmentation. Most of the architectures used in these works have a common feature extraction part (in the form of deep convolutional network) followed by two separate branches dedicated for the prediction of depth and semantic maps. In~\cite{cao2016exploiting}, it was found that the shared feature extraction part helps in making the semantic segmentation branch benefit from the learnt depth features in the shared encoder and leads to slightly better results in semantic segmentation. 

\textbf{Open problems:} None of the mentioned methods was benchmarked on aerial data, which incorporate very specif challenges in terms of variability of depth and semantic maps.
We benchmark our method and the joint RefineNet~\cite{nekrasov2019real} on MidAir dataset and we make the codes publicly available.
\section{Problem Statement}
\label{sec:statement}

We present the specifications of the problem we address. A monocular camera is rigidly attached to a UAV. The camera intrinsic parameters are assumed to be known and constant. The camera frame rate is relatively high such that there are overlapping regions between each two consecutive frames. The UAV moves freely (6 DoF) in an outdoor unstructured environment and records the video frames as well as the camera position (using a IMU sensor) at each time step. Using the position at each time step we can compute the motion transformation matrix \(T\) from one frame to the next.  Our objective is to design a network, denoted by a function \(F\), that takes at each time step the current frame \(I_{t}\), previous \(n\) frames \(I_{seq} = [I_{t-1}, I_{t-2}, ...,I_{t-n}]\) and camera motion transformations \(T_{seq} = [T_{t-1}, T_{t-2}, ..., T_{t-n}]\) and outputs an estimated depth map \(\hat{D_t}\) and semantic segmentation map \(\hat{S_t}\) corresponding to the current frame:

\begin{equation}
    (\hat{D_t}, \hat{S_t}) = F(I_t, I_{seq}, T_{seq}).
\end{equation}

The starting point of our work is the M4Depth network ~\cite{m4depth}. To the best of our knowledge, they produce the current top results in monocular depth estimation on the synthetic MidAir~\cite{midair} aerial dataset and their model weights and code are publicly available. In addition, the encoder-decoder modularity of their architecture makes it a convenient candidate to be transformed into a joint architecture. 


To this purpose, we first re-design the architecture to perform semantic segmentation instead of depth estimation. Then, we merge the two architectures into a single joint one, we call it Co-SemDepth, performing both depth estimation and semantic segmentation in real-time. 

\section{Methodology}
\label{sec:methodology}

In this section, we first give a brief overview of M4Depth for depth estimation, then describe our M4Semantic semantic segmentation network. Finally, we merge the two networks and flash light on our proposed joint Co-SemDepth architecture.

\subsection{Backbone Depth Network}

The architecture of M4Depth~\cite{m4depth} is an adaptation of the standard U-Net encoder-decoder network trained to predict parallax maps to be then transformed into depth maps. {The authors define parallax as a function of perceived motion, thus it can be seen as a general form of stereo disparity for an unconstrained camera baseline. 
}

The network takes as input a sequence of $n$ video frames (we choose \(n=3\)) and the camera transformation $T$ between each two consecutive frames. At each time step $t$ the encoder, same as in Figure~\ref{fig:joint}, takes a  $t$-frame $I_t$ and extracts image features at different scales using its pyramidal structure. Each encoder level is composed of two convolutional layers and a domain-invariant normalization layer (DINL)~\cite{dinl} to increase the network robustness to varied colors and luminosity conditions. 

Then, the decoder, see top part of Figure~\ref{fig:joint}, takes the feature maps at different resolutions obtained by the encoder at time $t$, the features extracted from the previous frame ($t-1$), the parallax map predicted at time $t-1$, and the camera motion transformation $T_t$ to predict the parallax map of the current frame $I_t$.  This parallax $p_t$ is then transformed into a depth map  $d_t$ following the transformation proposed in~\cite{m4depth}. 
Each level of the decoder is composed of a preprocessing unit and a parallax refiner. The preprocessing unit is responsible for preparing the input to the parallax refiner at this level and the parallax refiner is a stack of convolutional layers responsible for giving an estimate of the parallax map at each level.

 \textbf{Depth loss:} The network is trained in an end-to-end fashion and a scale-invariant \begin{math}L_1\end{math} loss is used to compute the loss between the predicted depth map \(\hat{d_i}\) and the ground truth one \(d_i\). The loss is computed at each decoder level and then accumulated across all the levels:

 \begin{equation}
     L_{depth} = \sum_{l=1}^{M}\frac{1}{N_p^l}\sum_{d_i^l} 2^{l+1}|log(d_{i})-log(\hat{d_{i}})|
 \end{equation}

 where \(M\) is the number of decoder levels and \(N_{p}^l\) is the total number of pixels in the image at level \(l\).

\subsection{M4Semantic Network}

\begin{figure}%
    \centering
    {\includegraphics[width=\linewidth, height=2.4cm]{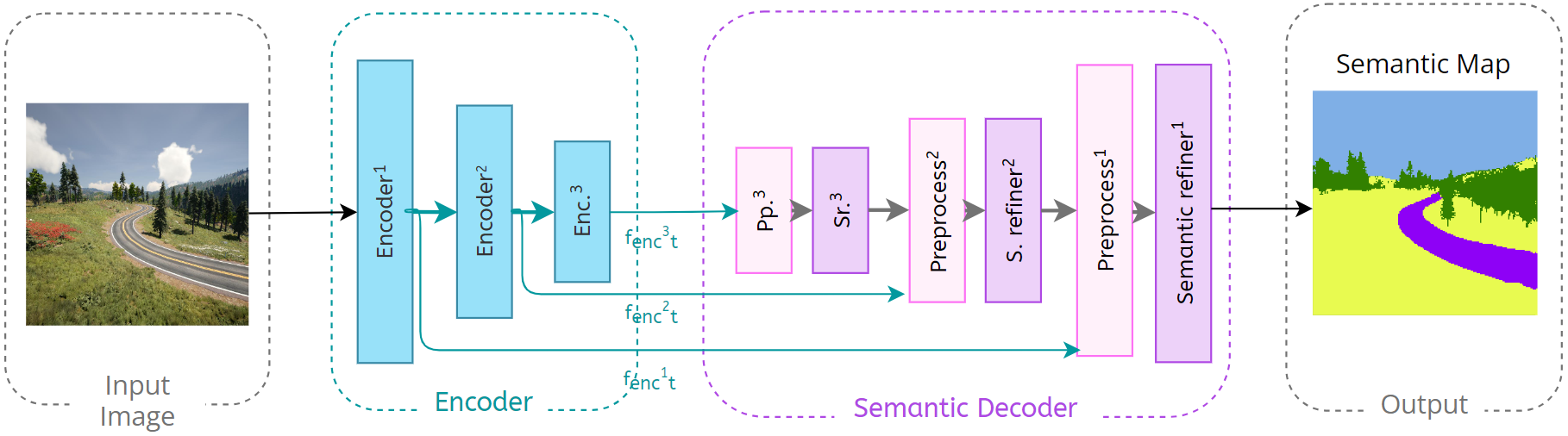} }
    \caption{\centering Our M4Semantic Architecture. It is composed of an encoder and a decoder module with a pyramidal structure. Each level of the decoder is composed of a preprocessing unit and a semantic refiner. 
    }%
    \label{fig:semantic}%
\end{figure}

Inspired by M4Depth, we propose a similar architecture for semantic segmentation depicted in Figure~\ref{fig:semantic}. 
The encoder is a stack of multiple levels and has the pyramidal structure where the resolution of the feature map is decreased while proceeding forward through the encoder levels. The feature map predicted at each encoder level is passed to its corresponding decoder level. Each decoder level is composed of a preprocessing unit and a semantic refiner. The preprocessing unit prepares the input to the semantic refiner and the semantic refiner at each level gives an estimate of the semantic segmentation map at a specific resolution. The resolution of the semantic map is scaled-up proceeding forward on the decoder levels.

\begin{figure*}%
    \centering
    {\includegraphics[width=0.7\linewidth, height=4.2cm]{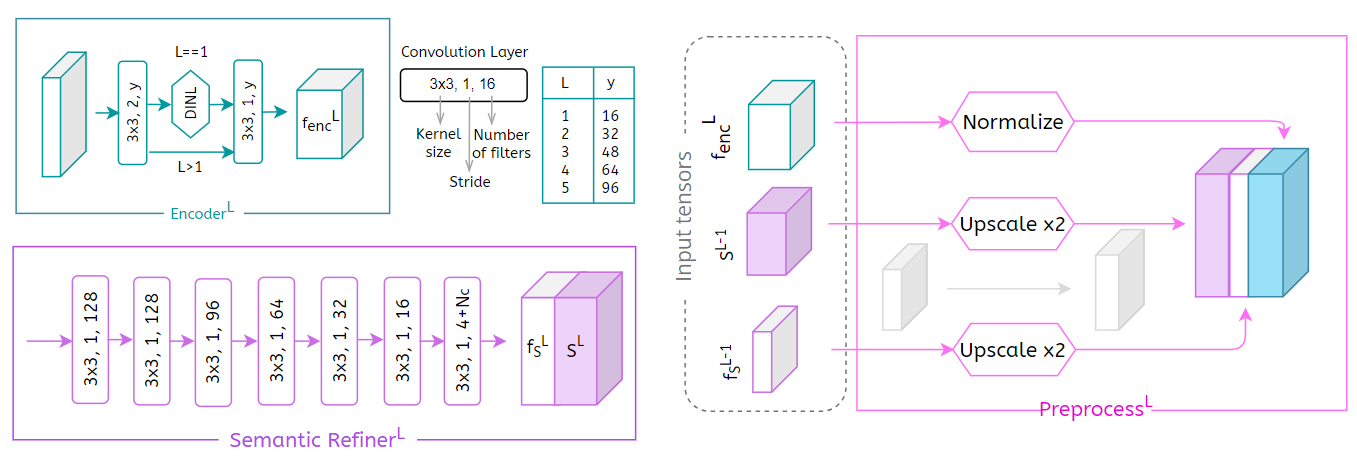} }
    \caption{\centering An illustration of the modules in our M4Semantic architecture. \(N_c\) is the number of semantic classes}%
    \label{fig:modules}%
\end{figure*}

 In Figure~\ref{fig:modules} we show the modules used in our architecture. The encoder at each level is composed of 2 convolutional layers. In the first level, DINL~\cite{dinl} is added after the first convolution to increase the network robustness to varied colors and luminosity conditions. ReLU activation is applied after each convolutional layer and the resolution is decreased by a factor of 2 after each level. The pyramidal structure of the encoder helps in extracting both coarse and fine (global and local) features from the input image.

The preprocessing unit at each decoder level is a pure computational unit with no parameters to be trained. It performs two operations:
\begin{itemize}
    \item It upscales the semantic map \(S^{L-1}\) and the semantic features \(f_{S}^{L-1}\) estimated from the semantic refiner of the previous level by a multiple of 2 to match the resolution of the current level.
    \item It normalizes the feature map \(f_{enc}^L\) received from the encoder.
\end{itemize}


Similar to the parallax refiner, the semantic refiner at each level is composed of a stack of convolutional layers. The last convolutional layer has a depth equals 4 (depth of the semantic features map) + N (the number of semantic classes). The output of the semantic refiner is a predicted semantic features map and an estimated semantic segmentation map. We apply Softmax activation on the predicted semantic segmentation map to obtain a probability score for each class on every pixel.

Different from M4Depth, our M4Semantic architecture works on single images. We removed the time dependency in the semantic decoder because this produced 2 times faster results than the one with time dependency with a slight drop in accuracy (this architectural choice will be discussed in Section V.E).   

\textbf{Semantic loss:} the standard categorical cross-entropy loss is used on the predicted semantic maps at each level. The ground truths are resized using Nearest Neighbour interpolation to match the resolution of the predicted semantic maps at intermediate levels. Then, these losses are aggregated through a weighted sum:

\begin{equation}
    L_{semantic} = \sum_{l=1}^{M}\frac{1}{N_p^l}\sum_{p_t^l} -log(\frac{p_{t}}{\sum_{j}^{N_c} p_{j}})
\end{equation}
where \(N_c\) is the number of semantic classes, \(p_{t}\) is the output softmax probability score for the target class and \(p_{j}\) is the output softmax probability score for class j.

\subsection{Joint Co-SemDepth Network}
To merge the two previously described networks, we adopt a multi-tasking shared encoder architecture~\cite{cao2016exploiting, mousavian2016joint, nekrasov2019real, zhang2018joint, he2021sosd}. The depth estimation and semantic segmentation networks share the encoder part for feature extraction, but each of them has its own decoder for their corresponding map prediction. An overview of our joint architecture is in Figure~\ref{fig:joint}. 

\textbf{Loss Function:} Our joint network is trained in an end-to-end fashion. The loss function for our architecture is defined as: 
\begin{equation}
    L_{total} = L_{depth} + w * L_{semantic}.
\end{equation}
We incorporated a weighting factor \begin{math}w\end{math} forcing the loss values for semantic to lie within the same range of the losses for depth, and thus ensuring a comparable contribution for the two losses during training.

\section{Experiments}
\label{sec:experiments}
In this section, we discuss the experiments we conducted to validate the effectiveness of our joint architecture. We first give a brief description of the datasets we used for training and evaluation; MidAir and Aeroscapes. Next, we report the implementation details. In the experimental analysis, we first evaluate the effectiveness of using our joint architecture compared to using the two single architectures. Then, we benchmark our model against other state-of-the-art methods. Finally, we make an architecture study.

\subsection{Datasets}
{\bf MidAir}~\cite{midair} is a synthetic dataset collected using AirSim simulator~\cite{airsim} consisting of 420K forward-view RGB video frames captured at low altitude in outdoor unstructured environments with various weather conditions. It contains annotations of depth maps, semantic segmentation, surface normals, stereo disparity, and camera locations. Hence, this dataset is suitable for training and testing our joint Co-SemDepth architecture.

We adopt the train-test split used in~\cite{m4depth}, but we randomly select 8 trajectories 
to create validation data. In the evaluation, the depth values are capped at 80.0 meters. We resized images to a resolution of 384x384 instead of the original 1024x1024. In the original semantic annotation of MidAir, there are 14 semantic classes: Sky, Animals, Trees, Dirt Ground, Ground Vegetation, Rocky Ground, Boulders, Empty, Water, Man-Made Construction, Road, Train Track, Road Sign, and Others. Since several classes are visually indistinguishable and some of them very small, we mapped them to a smaller set of 7 semantic classes : Sky, Water, Land, Trees, Boulders, Road, and Others. Specifically, we considered Ground Vegetation, Rocky Ground and Dirt Ground as {Land}, and Animals, Empty, Train Track and Road Sign in {Others}.

{\bf Aeroscapes}~\cite{aeroscapes} is a real dataset collected using drones at low-mid altitude in various outdoor environments. It consists of 3,269 images with 80\%-20\% train-test split and resolution of 1280x720. Some of these images are captured with a forward-view while others are captured with nadir (top) view. This dataset contains only semantic segmentation annotation. For this reason, we could not use this dataset for the training of our joint architecture; because it requires annotation of depth, semantic, and camera locations. However, we used Aeroscapes for the training and testing of our single M4Semantic network. There are 12 semantic classes in Aeroscapes, namely: Background, Person, Bike, Car, Drone, Boat, Animal, Obstacle, Construction, Vegetation, Road, and Sky.

\subsection{Implementation Details}
We adopt Adam optimizer with the default momentum parameters \((\beta_1 = 0.9, \beta_2 = 0.999)\) and a fixed learning rate of \(10^{-4}\). 
We apply image augmentation of random rotation, flipping, and changing color (contrast, brightness, hue, and saturation) during training and we train with batch size of 3. 
In the computation of the joint architecture loss $L_{total}$, after inspection of the loss ranges for depth (range from 0.0 to $\sim$1.0) and semantic (range from 0.0 to $\sim$7.0), we selected a weighting factor  \( w=0.1\). 
After training, we choose the checkpoint that produced the best validation results for evaluation on the test set.

Our workstation has 16GB RAM, Intel core i7 processor and a single NVIDIA Quadro P5000 GPU card running CUDA11.4 with CuDNN 7.6.5 and Ubuntu operating system.
Due to the memory limits of our workstation, we predict the depth and semantic maps at a resolution equal to half the input resolution and then apply Nearest Neighbour interpolation on the output maps to scale up its resolution to the original size. As reported in~\cite{chen2018driving}, decreasing the image resolution can slightly decrease the accuracy, however, it gives the advantage of reducing the computational runtime and memory footprint and these are critical factors for aerial robotics.

To quantitatively evaluate the depth prediction results, we consider the commonly used evaluation metrics in prior works~\cite{m4depth, manydepth, monodepth2}. These include the linear root mean square error (RMSE), the absolute relative error, and accuracy under a threshold. 
%
%
%
For semantic segmentation, we use the commonly used mean Intersection over Union \(mIoU\) metric. 
%
%
The Inference Time (Inf. Time) is computed in milliseconds per frame (ms/f).


\subsection{Joint vs Single Architecture}
We conduct experiments to compare the performance of our joint architecture Co-SemDepth to the two single architectures: M4Depth and M4Semantic, see Table~\ref{table_joint_vs_single}. Each architecture was trained equally for 60 epochs. 

We can notice that the accuracy values (in terms of depth and semantic metrics) of the joint architecture are close to the single ones. The inference time of the joint architecture (49.6 ms/f) is \emph{lower than} the sum of the two single ones (\(44.9 + 9.8\)). Moreover, the number of parameters of the joint architecture (5.2 Million) is less than the sum of the two single ones (3.06 + 2.61 Million) by around 500K parameters. During inference, Co-SemDepth required only \emph{6.2GB of GPU memory} while running M4Depth and M4Semantic together required 14.6GB of GPU memory. This makes Co-SemDepth compatible to run on the microcontroller Nvidia Jetson TX2 that has only 8GB RAM and that is widely used in robotics hardware.

The above signifies that using our joint architecture Co-SemDepth is more effective in terms of computational time and memory footprint than using the two single architectures while achieving very close accuracies. The trade-off between accuracy and computational cost in the multi-tasking architectures was previously discussed in~\cite{j7} and it can differ depending on the environment. 




\begin{table*}[h]
\begin{center}
\caption{\centering Evaluation of our joint vs single architectures for depth estimation and semantic segmentation on MidAir dataset. }
\label{table_joint_vs_single}
\resizebox{\linewidth}{!}{
\begin{tabular}{| c || c | c | c | c | c | c | c | c | c |}
\hline
 \multirow{2}{*}{Architecture}& \multirow{2}{*}{Output}  & \multirow{2}{*}{Params(M)} & \multirow{2}{*}{Inf. Time (ms/f)}   & Semantic Metrics & \multicolumn{5}{|c|}{Depth Metrics}\\ \cline{5-10}
  & & & & mIoU $\uparrow$ & RMSE $\downarrow$ & AbsRelErr $\downarrow$ & \(\delta < 1.25\) $\uparrow$ &  \(\delta < 1.25^2\) $\uparrow$ &  \(\delta < 1.25^3\) $\uparrow$\\
\hline
M4Depth & D & 3.06 & 44.9  & - & 6.821 & 0.0973 & 92.26\% & 95.62\% & 97.18\%\\ 
\hline
M4Semantic & S & 2.61 & 9.8 & 75.64\% & - &  - & - & - & -\\
\hline
\textbf{Co-SemDepth} & D+S &\textbf{ 5.2} & \textbf{49.6}  & \textbf{74.24}\% & \textbf{7.15} & \textbf{0.103} &  \textbf{92.1}\% &  \textbf{95.4}\% &  \textbf{96.94}\%\\
\hline 
\end{tabular}
}
\end{center}
\end{table*}

\subsection{Benchmarking}
{\bf MidAir: }We compare the performance of Co-SemDepth with other open-source state of the art methods (both single and joint architectures) in depth estimation and semantic segmentation, see Table~\ref{table_benchmark}. For each method, we fix the input size to 384x384 and the number of training epochs to 60. Other parameters for every method were kept as the default. More details about the parameters used for each baseline method can be found on our \hyperlink{https://github.com/Malga-Vision/M4Semantic/}{Github page}.

\begin{table*}[h]
\begin{center}
\caption{\centering Benchmarking our joint architecture on MidAir dataset against other state-of-the-art methods in both depth estimation and semantic segmentation. The top part reports single depth estimation methods, the middle part for single semantic segmentation methods, and the bottom part for joint methods. Methods marked with $^*$ means the depth metrics values were reported in~\cite{m4depth}.}
\label{table_benchmark}
\resizebox{\linewidth}{!}{
\begin{tabular}{| c || c | c | c | c | c | c | c | c | c |}
\hline
 \multirow{2}{*}{Method} & \multirow{2}{*}{Output} & \multirow{2}{*}{Params(M)} & \multirow{2}{*}{Inf. Time (ms/f)$\downarrow$}  & Semantic Metrics & \multicolumn{5}{|c|}{Depth Metrics}\\ \cline{5-10}
  & & & & mIoU $\uparrow$ & RMSE $\downarrow$ & AbsRelErr $\downarrow$ & \(\delta < 1.25\) $\uparrow$ &  \(\delta < 1.25^2\) $\uparrow$ &  \(\delta < 1.25^3\) $\uparrow$\\
\hline
MonoDepth2$^*$~\cite{monodepth2} & D & 14.8 & 23.9 & - & 12.351 & 0.394 & 61.0\% & 75.1\% & 83.3\%\\ 
ST-CLSTM$^*$~\cite{clstm} & D & 15.04 & 35.3 & - & 13.685 & 0.404 & 75.1\% & 86.5\% & 91.1\%\\ 
ManyDepth$^*$~\cite{manydepth} & D & 46.3 & 82.9 & - & 10.919 & 0.203 & 72.3\% & 87.6\% & 93.3\%\\ 
PWCDC-Net$^*$~\cite{pwc} & D & 9.4 & 25.8 & - & 8.351 & 0.095 & 88.7\% & 93.8\% & 96.2\%\\ 
\hline
FCN(VGG16)~\cite{fcn}  & S & 14.7 & 58.3 & 72.93\% & - & - & - & - & - \\ 
FCN(MobileNetv2)~\cite{fcn}  & S & 2.2 & 60.5 & 69.82\% & - & - & - & - & - \\ 
ERFNet~\cite{romera2017erfnet} & S & 2.07 & 19.1 & 77.4\% & - & - & - & - & - \\ 
SegFormer-B0~\cite{xie2021segformer} & S & 3.8 & 49.1 & 75.1\% & - & - & - & - & - \\ 
\hline
RefineNet~\cite{nekrasov2019real} & D+S & 3.0 & 74.2 & 72.7\% & 9.74 & 0.2 & 74.9\% &  89\% &  94.5\%\\ %
\textbf{Co-SemDepth (Ours)} & D+S & \textbf{ 5.2} & \textbf{49.6}  & \textbf{74.24}\% & \textbf{7.15} & \textbf{0.103} &  \textbf{92.1}\% &  \textbf{95.4}\% &  \textbf{96.94}\%\\ 
\hline
\end{tabular}
}
\end{center}
\end{table*}


We can clearly notice that our method outperforms the other joint network, RefineNet, in both semantic and depth accuracies and inference time. 

Compared to the single depth estimation networks, Co-SemDepth could maintain its superior accuracy in depth estimation. {This indicates that transforming M4Depth to the joint Co-SemDepth did not have a negative effect on its depth estimation performance compared to state-of-the-art}. Co-SemDepth also has a notable fewer number of parameters compared to the other methods. 

For the single semantic segmentation networks, Co-SemDepth has a competitive mIoU with the others, only slightly inferior to ERFNet, which is in any case a dedicated architecture. 

\begin{figure*}%
    \centering
    {\includegraphics[width=0.68\linewidth, height=9.6cm]{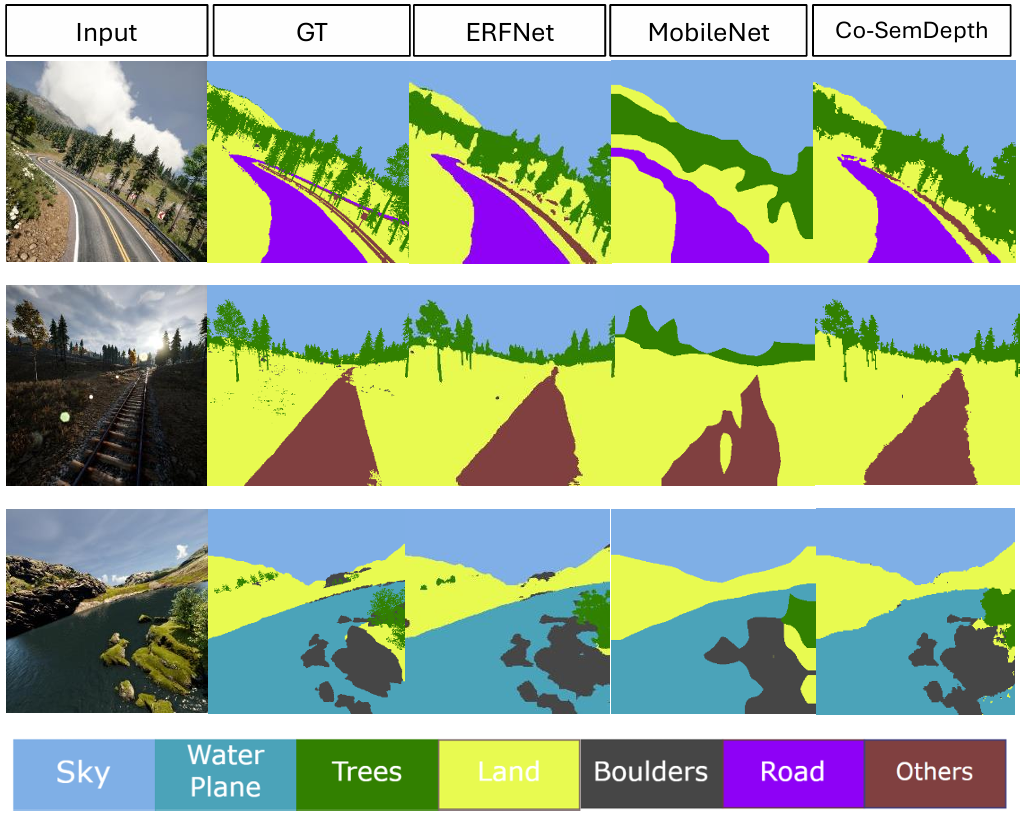} }
    \caption{\centering Qualitative evaluation of the semantic map predictions of ERFNet (3rd column), FCN MobileNet (4th column), and Co-SemDepth (5th column) on sample images from MidAir dataset.}%
    \label{fig:qualit}%
\end{figure*}

The per-class IoU evaluation can be found in Table~\ref{table_per_class} and a qualitative visualization of the semantic map predictions of three different methods (ERFNet, MobileNet, and Co-SemDepth) can be found in Figure~\ref{fig:qualit}. We can notice that the three methods could capture the overall semantic layout of the input images. However, Co-SemDepth and ERFNet are remarkably better in capturing the details (notice the trees and the train track) than MobileNet. Compared to Co-SemDepth, ERFNet is a bit better in capturing some of the faraway details (the distant boulders and bushes in the third row).

\begin{table*}[h]
\begin{center}
\caption{\centering Per-Class IoU Evaluation of Co-SemDepth architecture and other baseline methods on MidAir.}
\label{table_per_class}
\resizebox{\linewidth}{!}{
\begin{tabular}{| c || c | c | c | c | c | c | c || c |}
\hline
 Method & Sky & Water & Trees & Land &  Boulders & Road & Others & \textbf{mIoU} \\
\hline
FCN(VGG16)~\cite{fcn} &  88.56\% &  83.12\% &  75.5\% &  82.3\% &  29.97\% &  88.04\% &  54.6\% &  72.93\% \\
\hline
FCN(MobileNetv2)~\cite{fcn} &  87.82\% &  82.42\% &  73.42\% &  81.28\% &  26.57\% &  84.64\% &  46.09\% &  69.82\% \\
\hline
ERFNet~\cite{romera2017erfnet} &  91.5\% &  87.64\% &  82.48\% &  85.1\% &  40.63\% &  90.9\% &  63.7\% &  77.42\% \\
\hline
SegFormer-B0~\cite{xie2021segformer} &  90.54\% &  88.58\% &  79.7\% &  83.33\% &  30.13\% &  92.19\% &  61.2\% &  75.1\% \\
\hline
RefineNet~\cite{nekrasov2019real} &  89.7\% &  81.6\% &  79.7\% &  82.2\% &  30.15\% &  91.36\% &  54.1\% &  72.7\% \\
\hline
\textbf{Co-SemDepth} &  88.5\% &  83.7\% &  79.4\% &  81.7\% &  34.4\% &  94.7\% &  60.12\% &  74.24\%\\
\hline
\end{tabular}
}
\end{center}
\end{table*}

{\bf Aeroscapes: } We report the M4Semantic results in Table~\ref{table_aeroscapes}. Our network was trained for 200 epochs with a batch size of 3 and a learning rate of \(10^{-4}\) for the first 70 epochs and then decreased to \(10^{-5}\). 

\begin{figure*}%
    \centering
    {\includegraphics[width=0.67\linewidth, height=6cm]{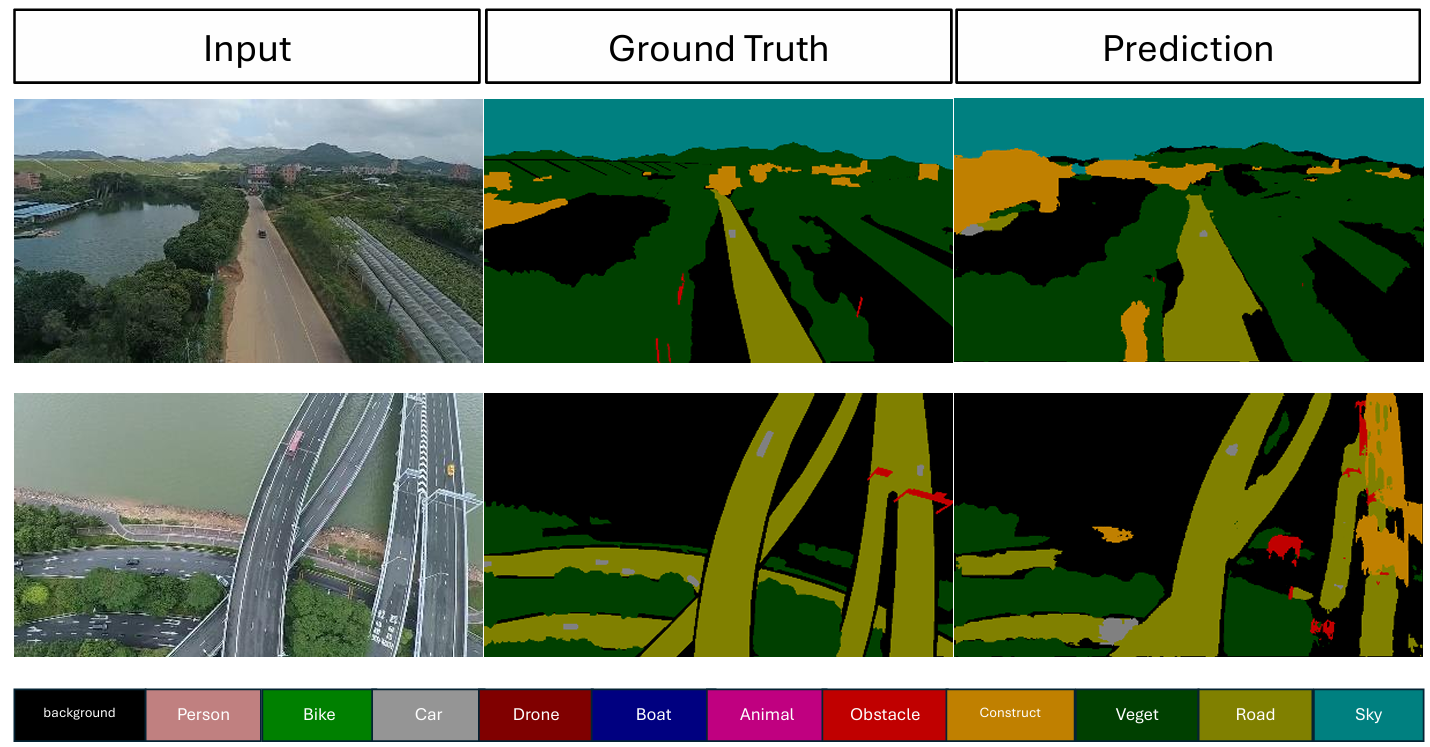} }
    \caption{\centering Visulaization of the predicted semantic maps using M4Semantic on Aeroscapes dataset.}%
    \label{fig:qualitaero}%
\end{figure*}

We implemented M4Semantic on TensorFlow as one whole model that can be trained in an end-to-end fashion without separation between the weight files of the encoder and the decoder. This led to a more compact code but limited us from pretraining the encoder separately on Imagenet as done in the other methods. 
Nevertheless, we could produce a competitive mIoU compared to the others.
An output visulization can be found in Figure~\ref{fig:qualitaero}. 

\begin{table}[h]
\begin{center}
\caption{\centering Comparison of our M4Semantic architecture with other semantic segmentation methods benchmarked on Aeroscapes. P means pretrained on other datasets and S means trained from scratch}
\label{table_aeroscapes}
\resizebox{\linewidth}{!}{
\begin{tabular}{| c || c | c | c |}
\hline
 Method  & P/S & Open-Source & mIoU $\uparrow$ \\
\hline
FCN-8S~\cite{zheng2023deep} & P & No & 43.12\% \\
\hline
FCN-16S~\cite{zheng2023deep} & P & No & 44.52\% \\
\hline
FCN-32S~\cite{zheng2023deep} & P & No & 45.51\% \\
\hline
FCN-ImageNet-4S~\cite{aeroscapes} & P & No & 48.96\% \\
\hline
FCN-Cityscapes~\cite{aeroscapes} & P & No & 49.55\% \\
\hline
FCN-ADE20K~\cite{aeroscapes} & P & No & 51.62\% \\
\hline
FCN-PASCAL~\cite{aeroscapes} & P & No & 52.02\% \\
\hline
\textbf{M4Semantic (Ours)}& S &\textbf{Yes} & \textbf{50.41}\%  \\
\hline

\hline
\end{tabular}
}
\end{center}
\end{table}


\subsection{Architecture Study}
We conduct architecture study experiments on M4Semantic to highlight the importance of the addition or ablation of different modules, see Table~\ref{table_ablation}. From the top part of the table, we can notice that using 5 levels produced the best mIoU. 

In the bottom part, Original means M4Semantic(5 level). In Original+\{SNCV\}, we used the Spatial Neighbourhood Cost Volume module used in the decoder in M4Depth~\cite{m4depth} on the encoded feature maps instead of adding the normalized feature maps directly in the preprocessing unit, Figure~\ref{fig:modules}. Such a module measures the two-dimensional spatial autocorrelation of the scene and improved the performance in depth estimation. However, as can be seen in the table, using such module in M4Semantic didn't improve the performance in semantic segmentation and, moreover, it increased the inference time.  

In Original+\{\(S^{t-1}\)\}, we test the addition of time dependency in M4Semantic. At each decoder level, the semantic segmentation map predicted of the previous frame \(S^{t-1}\)\ is used along with the camera motion information and the ground truth depth map to warp it and give an initial prediction of the semantic map of the current frame. We concatenate such warped map with the output of the preprocessing module, Figure~\ref{fig:modules}, at each decoder level. While such technique achieved higher mIoU, the inference time increased due to the added warping computation. Also, the warping of the semantic segmentation map requires the ground truth depth map and this is not available in most of the times in reality.

Given this architecture study and the one done on the single depth estimation architecture~\cite{m4depth}, we choose the number of levels of our joint architecture Co-SemDepth to 5 levels.

\begin{table}[h]
\begin{center}
\caption{\centering Evaluation of our M4Semantic architecture on MidAir with the addition (+) or ablation (-) of different modules. The top part evaluates choosing different number of levels. The bottom part was performed on M4Semantic(5level).}
\label{table_ablation}
\resizebox{\linewidth}{!}{
\begin{tabular}{| c || c | c |}
\hline
 Architecture &   Inf. Time (ms/f) $\downarrow$  & mIoU $\uparrow$ \\
\hline
M4Semantic(4level) & 9 & 73.93\% \\ 
\hline
M4Semantic(5level) & 9.8 & 75.64\% \\ 
\hline
M4Semantic(6level) & 10.9 & 73.7\% \\ 
\hline
\hline
Original-\{DINL\} & 9.5 & 74.42\% \\ 
\hline
Original-\{Normalize\} & 9.7 & 72.94\% \\ 
\hline
Original+\{SNCV\} & 17.7 & 70.65\% \\ 
\hline
Original+\{\(S^{t-1}\)\} & 16.7 & 77.2\% \\ 
\hline
\end{tabular}
}
\end{center}
\end{table}


\section{Conclusion}
\label{sec:conclusion}

In this work, we presented Co-SemDepth, a fast joint architecture for depth estimation and semantic segmentation for aerial robots. 
Our architecture proved to be light and fast compared to the other methods while achieving better or on par accuracies. This makes it very suitable to be deployed on UAV hardware for real-time scene analysis in outdoor environments. 
Using our light architecture makes it possible for the drone to conduct scene analysis autonomously and independently using its onboard microcontroller without the need for sending data to a remote server to carry out the analysis. The output depth and semantic maps can provide geometric and semantic understanding of the scene that is necessary to carry out a variety of UAV missions.


\bibliographystyle{ieeetr}
\bibliography{references}          
\nocite{*}

\end{document}